\documentclass[11pt]{article}
\usepackage{graphicx}
\usepackage{listings}

\title{\bf A Semi-Automatic Framework to Discover Epistemic Modalities in Scientific Articles}
{\author{\small Sviatlana Danilava\\
             \small JW Goethe-University Frankfurt am Main\\
             \small Dept. of Computer Science and Mathematics\\
             \small Robert-Mayer-Str. 11-15, D-60486 Frankfurt am Main, Germany.\\
             \small Email: danilava@cs.uni-frankfurt.de
             \and
             \small Christoph Schommer \\ 
              \small University of Luxembourg\\ 
              \small Dept. of Computer Science - ILIAS Laboratory, MINE Research Group\\
              \small 6, Rue Richard Coudenhove-Kalergi, 1359 Luxembourg, Luxembourg\\
              \small Email: christoph.schommer @ uni.lu Home: mine.uni.lu
}
\date{\today}

\markboth{A}{B}

\begin{document}
\maketitle

\begin{abstract}
Documents in scientific newspapers are often marked by attitudes and opinions of the author and/or other persons, who contribute with objective and subjective statements and arguments as well. In this respect, the attitude is often accomplished by a linguistic modality. As in languages like english, french and german, the modality is expressed by special verbs like {\sf can, must, may, etc.} and the subjunctive mood, an occurrence of modalities often induces that these verbs take over the role of {\it modality}. This is not correct as it is proven that modality is the instrument of the whole sentence where both the adverbs, modal particles, punctuation marks, and the intonation of a sentence contribute. Often, a combination of all these instruments are necessary to express a modality. In this work, we concern with the finding of modal verbs in scientific texts as a pre-step towards the discovery of the attitude of an author. Whereas the input will be an arbitrary text, the output consists of zones representing modalities.
\end{abstract}

\section{Introduction}
Search engines that base on the World-wide Web find large amounts of hits and information by any request. However, intelligent search queries like {\sf Which scientists support hypothesis A} or {\sf Does the author believe in my opinion} are not yet supported. In order to answer, a search engine must  first search for appropriate documents and then analyse them fast. For this, intelligent algorithms are required that take into account linguistic insights for a analytical consideration of syntax and style but also for a treatment with meta-aspects like opinion and attitude of the author himself. This is especially true for scientific texts: they are very objective in concern of the description of a hypothesis or in discussing diverse problems. The discovery of the subjective opinion or attitude of the author is a major topic and is the research objective of \textit{Attitude Mining}. It concerns with the discovery of meta-information out of documents, especially the attitude of an author in respect to events, references to other's work, etc. The attitude can be positive or negative, but in most cases, it is hidden and to be proved by indications. Attitude Mining concerns with the explorative discovery of these indications (\cite{SQW06}), but demands for profound knowledge in areas like computer science, linguistics, cognitive sciences and psychology. Following \cite{HOL88}, there exist more than 350 lexical style attributes for the attitude, for example to express doubts or beliefs. To further motivate, the following sentences should demonstrate the existence of subjectivity in scientific texts:

\begin{itemize}
    \item[$\rightarrow$] {\sf When paleontologists seek the roots of life, they head to rocks of the Archaean Eon, which range from 3.8 billion to 2.5 billion years old.} 
    \item[$\rightarrow$] {\sf Australian and Canadian researchers argue this week in Nature that stromatolites were so diverse and complex that they must have been alive.}
    \item[$\rightarrow$] {\sf Martin Brasier of Oxford University is less sanguine, arguing that the structures are more likely chemical precipitates. He also objects to the reasoning in the Nature paper. ``You can‘t use the argument that complexity is the signature for life,'' he says. }
\end{itemize}

The first sentence is neutral, as it describes only a procedure what palaeontologists normally do {\it when they try out to find the origin of life}. The second sentence holds a hypothesis with explanatory statements, the third sentence arguments against the hypothesis in the second sentence having the author/originator referenced. In this respect, the modality concerns with the speaker's style to modify the proposition of sentences through subjective components. And as we have seen above, many sentences are modal, for example

\begin{itemize}
    \item[$\rightarrow$]  {\sf I believe she arrives this morning at London Heathrow.}
    \item[$\rightarrow$]  {\sf I can not be in today.}
\end{itemize}

In western languages, the modality is expressed by special verbs like {\sf can, must, may, etc.} and by the subjunctive mood. However, this often induces that verbs take over the role of {\it modality}, which is not correct: it is proven that modality is the attribute of the whole sentence where both the adverbs, modal particles,  punctuation marks, and the intonation of a sentence contribute to it. Often, a combination of all these instruments are necessary to express a certain modality. For example, the sentence

\begin{itemize}
     \item[$\rightarrow$]  {\sf Do you really think that?}
\end{itemize}

leads to another understanding as with

\begin{itemize}
     \item[$\rightarrow$]  {\sf You do not really think of that?}
\end{itemize}

In the first sentence, the combination of \textit{really}, \textit{think} and the transfer into a question is very subjective, but leaves the recipient some space. However, the second sentence is much more subjective, influencing the recipient's answer completely and leaving no space for another answer than 'no'. Overall, the complexity in using modalities is one of the major problems, both for the analysis of texts per se and for machine translation systems. 

In this work, we concern with finding modal verbs in scientific texts as a pre-step toward discovering the attitude of an author. Whereas the input will be an arbitrary text, the output consists of zones representing modalities.

\section{Fundamentals}
Originally, the concept of modality derives from the formal logic. Here, a modal expression consists of two parts, the \textit{modal} part and the \textit{proposition} part. The modal part contains the modality, the proposition the actual statement. Moreover, the modal part is either \textit{deontic} or \textit{epistemic} (\cite{PAL01}). A deontic modality describes the conditions that leads the statement to true or false, always being in relation with the reality, for example:

\begin{itemize}
    \item[$\rightarrow$] {\sf Indeed, the turnover of phytoplankton can be so high that there can be inverted pyramids of biomass, in which the standing crop of herbivorous zooplankton actually exceeds that of the phytoplankton. }
\end{itemize}

The verb acts as modality, it expresses a certain idea of the objective reality that might come true under certain circumstances. The epistemic modality, on the other side, concerns with personal experiences and a knowledge level of the author, but less with reality:

\begin{itemize}
    \item[$\rightarrow$] {\sf Australian and Canadian researchers argue this week in Nature that stromatolites were so diverse and complex that they must have been alive.}
\end{itemize}

The verb \textit{must} appears in an epistemic way, the statement is not proven yet but still an assumption. This assumption is proven initiated by a justification. Furthermore, the source of information is given, for example in

\begin{itemize}
    \item[$\rightarrow$] {\sf Martin Brasier of Oxford University is less sanguine, arguing that the structures are more likely chemical precipitates. }
\end{itemize}

This sentence contains an explicit source, namely \textit{Martin Brasier}. Such statements are referenced as evidential statements and are mostly referenced as a sub-category of an epistemic modality.

The modality is supported by a set of expressions: in order to develop a methodology in respect to an automatic recognition, the lexical fundament must be found first. Modal verbs form a class of verbs that add a modal meaning to a proposition. They allow the sender to modify the essence of a sentence by possibilities, necessities, doubts, beliefs, etc. In the English language, this is for example {\sf must - have to}, {\sf can - could - may}, and {\sf will - would - shall}. However, the use of modal verbs often leads to ambiguity as the same modal verbs are taken to express both the deontic and the epistemic relevance. Verbs like {\sf believe, doubt, accept, reject, etc.} describe the mental state of the speaker or his attitude against propositional part of the statement. Moreover, \textit{noun} may describe the mental states or cognitive processes as well, for example by \textit{doubt, belief, rejection, etc.}. Adverbs and adjective are \textit{lexical modifiers} that may assign doubts and beliefs, for example \textit{perhaps, probably, possibly, certain, likely}, etc.

English modal verbs are used both in epistemic and in deontic meanings. Generally, modal verbs express either a possibility or a necessity; each modal verb offers several meanings with semantic and pragmatic differences, for example the word \textit{must}. In the deontic version, it describes a necessity with the consideration of an external source, where the propositional subject is not source of modality. In contrast to this, an epistemic version describes a necessity, taking a logical justification. The following two sentences are deontic (first) and epistemic (second):

\begin{itemize}
    \item[$\rightarrow$] {\sf I must go, she is already waiting for me.}
    \item[$\rightarrow$] {\sf Where is John? It is 14h00, he must be in school.}
\end{itemize}

The epistemic reading of modal verbs can be summarised as follows:

\begin{itemize}
    \item Epistemic necessity as a conclusion out of the speaker's evidence: \textit{she must be in her office}.
    \item Epistemic necessity as logical conclusion out of a common valid and known fact:  \textit{she will be in her office}.
    \item Epistemic possibility as an uncertainty of the speaker:  \textit{she may be in her office}.
\end{itemize}

The epistemic usage of modal verbs, the epistemic adverbs and cognitive verbs distribute the subjectivity. The provide a basis for the attitude of the author, as for example in

\begin{itemize}
    \item[$\rightarrow$]{\sf The individual grains in them could not have accumulated mechanically because the slope of the cone is too great,“ says Stanley Awramik, a stromatolite expert at the University of California, Santa Barbara, who was not involved in the research.}
\end{itemize}

Here, the proposition is just a personal attitude (\textit{could}) of the referenced person, that is not proven at all. Given by the modal verbs, there is still enough information to discover the author's attitude and to differentiate the author's attitude against others' attitudes. 

\section{Selected Research Work}

The current research follows divergent directions, especially in the establishment of linguistic and cognitive models. These models support an understanding of the lexical means of expression, their influence to the lexical environment, and the modification of meaning while using modality. 

In respect to modalities as a influencing component to discovering the attitude, \cite{PZ06} says that it is insufficient to implement the attitude as to be positive or negative. Moreover, the attitude can be modified via {\it contextual valence shifters} by {\sf not}, {\sf never}, {\sf none}, but must take into account modifiers like {\sf rather}, {\sf deeply}, and/or {\sf few}. \cite{BER06} says that a {\it reported speech} shares a particular attention, since evidential aspects must be examined additionally. \cite{KEF06} argues that the lexical means of expression should not become considered as conveyor of meaning, but typical structures of attitude phrases can be observed.

The analysis of lexical resources that is additionally used to highlight the intention of the authors to produce attitudes is currently under research as well. \cite{MAT06} follows an establishment of specific emotional lexicons with positive, negative, and neutral meaning as well as an automatic extraction of emotion to extend these lexicons. 

The detection of document zones to structure the document becomes more and more popular. Initially, it has been presented as \textit{Argumentative Zoning} by Teufel and Moens  (\cite{TM99}), but has been applied in other works as well (\cite{ST07}, \cite{TEU06}) or strongly influenced research work on \textit{Content Zoning} (\cite{BRU08}). The main motivation is to \textit{summarise} documents and to zone in discourse-rhetoric zones. Teufel and Moens argue that - depending on the type, genre and style of the text - a standardised structure can often be identified. Using \textit{scientific articles}, they have assigned seven argumentative zones to each text, the zoning is then performed by a supervised learning system. \cite{MMC04} suggests an extend classification where each sentence is assigned to a rhetoric role. There exist up to ten zones that are classified into 3 classes. They argue that there exist no sequences of rhetoric roles; sentence may belong to different zones, also called as combined zones.

Following the idea of \textit{Opinion Mining}, \cite{BYT06} describe a model to detect \textit{opinion words}. The idea is to discover propositions, which contain subjective lexical expressions and the proposition itself, for example in combination with \textit{accuse}, \textit{criticise}, or \textit{doubt}. All constituents of each sentence receive a zoning label like \textit{Opinion Proposition}, \textit{Opinion Holder} or \textit{Null}. Another approach are disambiguation processes of modal verbs, where \cite{KIP95} has implemented a rule-based system towards the disambiguation of the epistemic and deontic meaning of the german verbs like {\sf sollen}, {\sf k\"onnen}, or {\sf d\"urfen}.

\section{Architecture}

\begin{figure}[htbp]
   \centering
   \includegraphics[width=7cm]{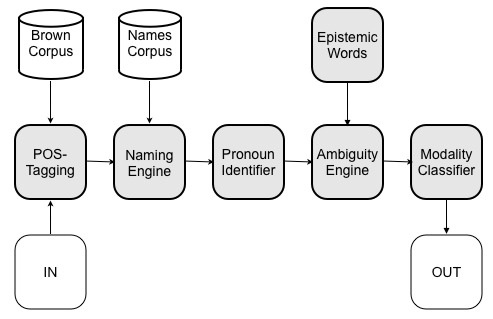} 
   \caption{The architecture of the framework, using the Brown Corpus and the Names Corpus. First, a part-of-speech tagger\cite{BLK07} to a given input and sends the intermediate result to the \textit{naming} and \textit{pronoun engine}. After that, modalities are disambiguated (\textit{Ambiguity Engine}) while synchronised with the list of modality verbs and finally classified (Modality Classifier). The text output is contains \textit{modality tags}.}
   \label{fig:architecture}
\end{figure}

The framework of this work consists of two major parts which are presented in Figure \ref{fig:architecture}. The first part focus on pre-processing the input text whereas the second part concerns with the disambiguation of the modal verbs and semi-automatic classification of the corresponding sentences. The pre-processing begins with a part-of-speech tagger, and is followed by a module to detect the naming entities and the pronouns. 

\vskip 0.5cm

We have taken an advantage in a way that we have used the WordNet thesaurus (\cite{FEL98}) to establish a list of modality verbs. This list contains several lexical categories and are found recursively by using a synonym function of WortNet. Currently, there exist several corpora for the English language, for example the Brown University Corpus (\cite{KUF67}), the International Corpus of English (ICE), and the British National Corpus (BNC). The ICE is a set of corpora that supports various dialects of English from around the world. The BNC is a text corpus with both written and spoken English words, covering covers more than 100 million words of the late twentieth century from a wide variety of genres. However, in this work, we use the Brown University Corpus to support the part-of-speech tagger to assign a syntactic category to each word of the input document. Here, the word is kept as it occurs, meaning that the original word is substituted by a list of syntactic categories and the original word. Words of the same root but of another flexion are kept as they are.

\begin{figure}[htbp]
   \centering
   \includegraphics[width=12cm]{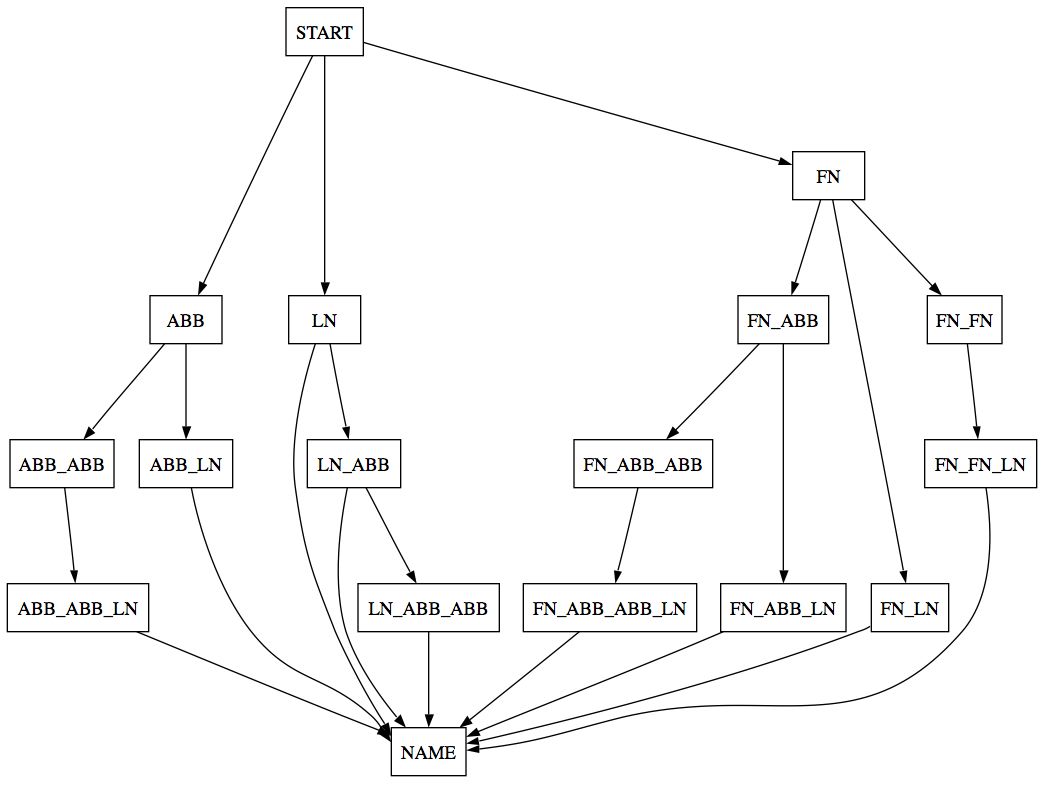}
   \caption{Automaton used for Naming Detection where \textit{FN} corresponds to the full first name, \textit{LN} to the full last name, and \textit{ABB} to any kind of abbreviations, like the abbreviated middle name. For example, P. Green follows the path of \textit{ABB\_LN}, whereas \textit{Peter Green} empties in \textit{FN\_LN}.}
   \label{fig:graph}
\end{figure}

In concern of dissolving \textit{naming entities}, a first method concerns with identification of personal names on the basis of references that are probably given in the document. Per definitionem, this method is firstly applied but suffers from diverse proper names of institution names like {\sf Max-Planck Institute}. In this case, external databases must be consulted using an automaton (see Figure \ref{fig:graph}). For the identification of person names, we have used the \textit{Names Corpus} by \cite{KRR91}, which contains 5001 female and 3000 male first names.

To identify the pronouns in the text, we restrict the list of possible candidates and consider only \textit{he}, \textit{she}, and \textit{who} as they concretely reference to one specific person. Common terms like \textit{researchers} or \textit{community members} are not considered as well as the pronoun \textit{they} and \textit{cataphora}. To identify the pronouns, we firstly concern with \textit{who}, which occurs after a referenced nominal phrase (NP) but in the same sentence as a NP.

After having pre-processed the data, the annotated texts are then sent to the classification module. The modal verbs are first disambiguated before they are sent to the classifier. As we must differentiate between deontic and epistemic modality, these two classes are taken as classes. We then use the following simple rule scheme:

\begin{itemize}
    \item A sentence is \textbf{deontic modal} if it contains a modality word that is deontic and if there exist modality words, which reference to facts.
    \item A sentence is \textbf{epistemic modal} if it contains a modality word that is epistemic and there exist modality words, which reference to subjective attitude of the author.
    \item A sentence is \textbf{non-modal} if there is no lexical evidence for modality.
\end{itemize}

\begin{figure}[htbp]
   \centering
   \includegraphics[width=12cm]{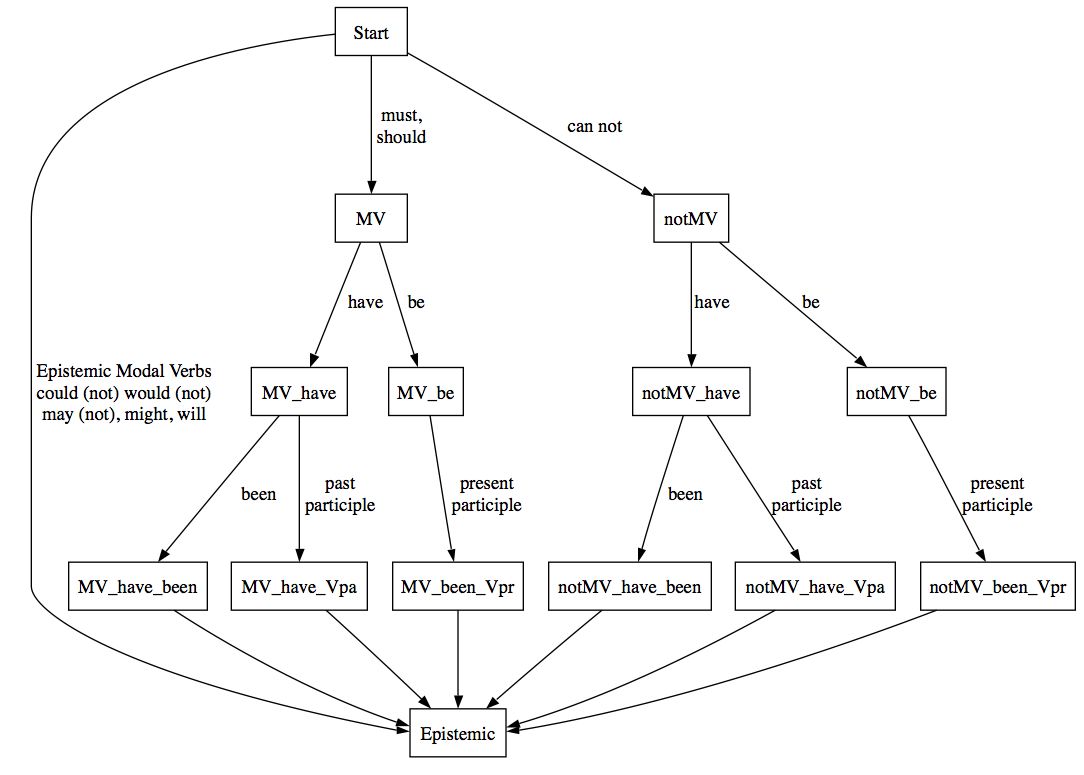}
   \caption{Automaton used for disambiguation where \textit{MV} corresponds to modal verb, \textit{Vpa} the verb past participle, and \textit{Vpr} the verb present participle. \textit{not} represents the negation, \textit{have}, \textit{be}, and \textit{been} the corresponding words.}
   \label{fig:graph2}
\end{figure}

The scheme may become improved when other criteria for disambiguation are included, for example the time. A more granular differentiation between \textit{epistemic positive} and \textit{epistemic negative} is possible when considering together the modal and propositional part of the sentence and classifying the sentences into \textit{Author X believes in Y} (positive) and \textit{Author rejects Y} (negative). The disambiguation process is shown as disambiguation automaton in Figure \ref{fig:graph2}.

The automaton decides to which class a modal verb belongs to. Depending on certain collocations, the a-priori probability for a modal verb to be epistemic is generally higher than to be deontic, so that we take a decision quite early. For example, if a certain collocation proves that a modal verb $v_i$ is probably epistemic for 90 percent, the automaton classifies $v_i$ as to be epistemic. The classification criteria are:

\begin{itemize}
   \item The modal verb {\sf must} refers to an epistemic necessity if it occurs with the following components: {\sf have been}, {\sf be} and {\sf verb present participle}, {\sf have} and {\sf verb past participle}, {\sf have been} and {\sf verb present participle}. In all other cases, the verb should be deontic.
   \item {\sf can} is deontic.
   \item {\sf can not} refers to the same verb components than {\sf must}.
   \item {\sf could} is epistemic.
   \item {\sf may} is epistemic.
   \item {\sf might} is epistemic as a tentative version of {\sf may}.
   \item {\sf will} is epistemic since future aspects are still hypothetic.
   \item {\sf shall} is deontic.
   \item {\sf should} is epistemic as {\sf must}. 
\end{itemize}

In Figure \ref{fig:graph2}, only the paths to the class epistemic are shown; it is assumed that all other paths are either deontic or non-modal. A path like

\begin{center} {\sf MV$\rightarrow$have$\rightarrow$been} \end{center}

means that the sentence contains a sequence of verb and \textit{have} and \textit{been}. Modal verbs express the attitude and opinion of a person. In this work, we concern with two types of persons: 

\begin{itemize}
    \item The person is the author: the author of the text gives an opinion and attitude about hypotheses, other authors, or other methods. This often occurs in scientific articles, for example references or citations. In the following example, the attitude is expressed by the author himself:
     \begin{itemize}
          \item[{$\rightarrow$}] {\sf Would a 100 mm scanning resolution be sufficient to produce an accurate model for paleontological study, or is a 50 mm scanning resolution a requirement?}
     \end{itemize}
     \item The person is the third person: this happens if the author speaks about other persons and presents those attitudes. In this case, these persons are referenced explicitly by name or work. This is a typical way of discussions in scientific articles. 
     \begin{itemize}
          \item[{$\rightarrow$}] {\sf Lowe pointed out their resemblance to modern forms but later had doubts.}
     \end{itemize}
\end{itemize}

To estimate the attitude of an author, we only consider epistemic sentences. We mark the importance of the modal part by a predicate $M$ and $\neg M$, respectively, and the propositional part by a predicate $H$, if the propositional part contains arguments pro $M$, otherwise $\neg H$. All epistemic sentences can be described with

\begin{center}
   $M(H)$ or $M(\neg H)$ or $\neg M(H)$ or $\neg M(\neg H)$
\end{center}

However, this step can become conditionally automated as it is quite hard to decide if the modal part is $M$ or $\neg M$: to do so, we certainly must find out the lexical information about a modal verb inside its lexical environment. Some modifiers like {\sf less} or {\sf more} and negations like {\sf not} or {\sf none} must be taken into account as they modify the meaning of the modal verbs. Their scope is important; an exact analysis implies the definition of complex grammars. Secondly, we must decide if the propositional part is $H$ or $\neg H$, so that we concern with propositional content analyis. This could be done with thesauri like WordNet, as these contain descriptions of relationships between words, for example synonyms. For example, WordNet allows a multiple calculation of similarity between words, depending on the distance between these words in the thesaurus: the shorter the distance, the similar the words.

In this work, we have identified two problems: first, the similarity between two words does not correspond to the actual situation in the text and second, the similarity can only be computed between pairs of words, but not between phrases or sub-phrases. We may say at this point that the architecture is \textit{hybrid}, meaning that the last step of estimating the attitude is done manually - based on the result that is produced. We then finally get a text result that is composed of text and meta information, consisting of two parts: the first part is machine readable as the data structures stay constantly with tags and structural information; it can therefore further be processed. The second part contains the epistemic sentences. A third and last step concerns with the segmentation of epistemic sentences depending on the hypothesis of the text. For this, we may use a graph, where all referenced persons are classified into three classes: \textit{Pro} references all members $P$, \textit{Contra} all members $C$, and \textit{Neutral} all members $N$. Each group can be empty, but not at the same time, as the author must belong to at least one class. We then assign

\begin{itemize}
   \item \textit{Pro} refers to sentences of $M(H)$ and $\neg M(\neg H)$
   \item \textit{Contra} to sentences of $M(\neg H)$ and $\neg M(H)$
   \item \textit{Neutral} collects undecidable sentences, especially of those persons who decline a decision.
\end{itemize}

\section{Example}
The following steps show an example using the following scientific text:

\begin{itemize}
     \item[$\rightarrow$] {\sf "The individual grains in them could not have accumulated mechanically because the slope of the cone is too great," says Stanley Awramik, a stromatolite expert at the University of California, Santa Barbara, who was not involved in the research.}
\end{itemize}

Generally, figures, formulas, and charts are manually pre-processed and substituted by tag-placeholders like \textit{FIG} or \textit{MATH}. The part-of-speech tagger then marks the text by two subsequent loops, where first all words are matched against the \textit{Brown Corpus}. Often, domain-specific termini arise, which are unknown and therefore labeled by a \textit{None}. Therefore, a second loop takes into account the morphologic structure of these words, for example, assigning a suffix \textit{tion} to the category \textit{noun}:

\begin{itemize}
     \item[$\rightarrow$] {\sf [(The, ART), (individual, ADJ), (grains, NNS), (in, IN), (them, PPO), (could, MV), (not, *), (have, HAVE), (accumulated, VPA), (’mechanically’, ’RB’),...] }
\end{itemize}

Recognizing the names, we then check if the text contains a list of references: in the positive case, all names are marked by a \textit{Person}-tag. These words that begin with an uppercase letter are considered as well and set to candidates of possible first and last names, abbreviations, or other personal names. They are marked by a \textit{NP}-tag. The first names are matched up with the mentioned \textit{Names Corpus}. However, as ambiguity may occur, such words are disambiguated manually. We then get the following automaton as it has been described in Figure \ref{fig:graph2}:

\begin{itemize}
     \item[$\rightarrow$] {\sf \dots $<$Person$>$ (Stanley, NP) (Awramik, NP) $<$/Person$>$, \dots}
\end{itemize}

The decision, to which objets a personal pronoun belongs to, is taken by considering the lexical categories \textit{PPS} and \textit{WPS}:

\begin{itemize}
     \item[$\rightarrow$] {\sf \dots $<$Person$>$ (Stanley, NP) (Awramik, NP) $<$/Person$>$, \dots$\\\dots <$Person Name= Awramik$>$ (who, WPS) $<$/Person Name= Awramik$>$}
\end{itemize}

The final classification then leads us to 

\begin{itemize}
     \item[$\rightarrow$] {\sf $<$EPISTEMIC$>$\\ \dots (could, MV), (not, *), (have, HAVE), (accumulated, VPA), \dots\\ $<$/EPISTEMIC$>$ }
\end{itemize}

where the tag \textit{EPISTEMIC}, \textit{DEONTIC}, or \textit{NON-MODAL} represent the modal state. As modelled in Figure \ref{fig:graph2}, the phrase {\sf could not $\rightarrow$ have $\rightarrow$ accumulated} is ambiguous and leads to \textit{negMV\_HAVE\_VPA}. The sentence therefore is marked as epistemic.

\section{Classification Results}
\begin{figure}[htbp] 
   \centering
   \includegraphics[width=12cm]{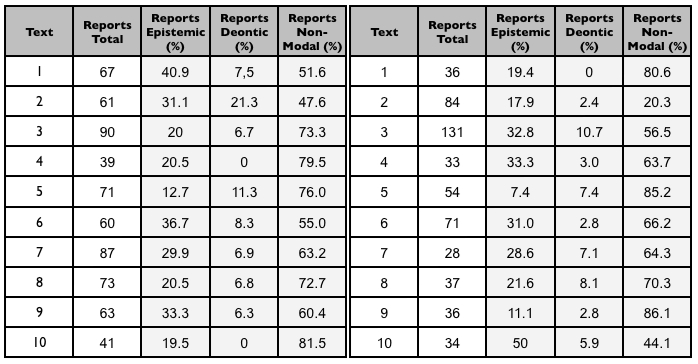} 
   \caption{Percental distribution of epistemic, deontic, and non-modal sentences, where the left chart corresponds to {\sf SCA}, the right chart to {\sf SCI}.}
   \label{fig:result1}
\end{figure}

We have used scientific articles from the fields of palaeontology and biology as a first test set (in the following called {\sf SCA}) and contributions to the scientific newspaper (in the following called {\sf SCI}) as a second test set. All test documents of {\sf SCA} share a common frame like \textit{Author X talks about his work Y}; text documents of {\sf SCI} share a frame like \textit{Author X talks about the opinions of M scientists in respect to hypothesis Y}. For {\sf SCA}, the texts share a similar length and style; the number of epistemic sentences is dominant to deontic and/or non-modal sentences (see Figure \ref{fig:result1}).

\begin{figure}[htbp] 
   \centering
   \includegraphics[width=12cm]{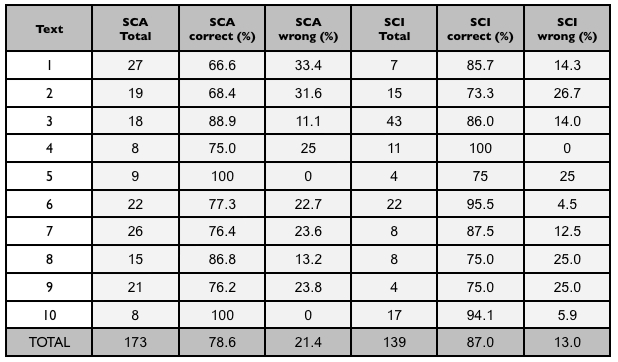} 
   \caption{Percental classification result of selected sentences of {\sf SCA} and {\sf SCI}. The correct classified sentences are higher for {\sf SCI} (87\%) than to {\sf SCA} (78.6\%). }
   \label{fig:result2}
\end{figure}

In total, 312 sentences have been analysed where 55.4\% are of {\sf SCA} and 44.6\% from {\sf SCI}. As presented in Figure \ref{fig:result2}, the correct classified sentences for {\sf SCI} (87\%) are higher than to {\sf SCA} (78.6\%). In respect to the wrong classified sentences, the modal word {\sf will} occurs most frequently. The following list shows some epistemic sentences that are classified correctly and wrongly:

\begin{itemize}
    \item[$\rightarrow$]{\sf EPISTEMIC This evidence of an ecological shift preceding phenotypic change suggests that this part of the sequence {\bf may} record rapid evolution driven by shifts in trophic ecology and adaptation to benthic niches.(correct)}
    \item[$\rightarrow$]{\sf EPISTEMIC If this {\bf hypothesis} is correct however the low number of specimens displaying intermediate phenotypes is puzzling and the scenario of replacement of one lineage by another cannot be ruled out. (correct)}
    \item[$\rightarrow$]{\sf EPISTEMIC Yet direct evidence that feeding controls evolution over extended time scales available only from the fossil record is difficult to obtain because it is rarely {\bf possible} to directly analyze dietary change in long-dead animals. (wrong)}
    \item[$\rightarrow$]{\sf EPISTEMIC First {\bf perhaps} the best-known work on specialisation in fishes concerns stickleback in postglacial coastal lakes in Canada where planktivores and benthic feeders coexist as two reproductively isolated and phenotypical distinct tropic.(wrong)}
    \item[$\rightarrow$]{\sf EPISTEMIC Laboratory feeding experiments and analyses of wild stickleback populations {\bf show} that micro-wear exhibits a progressive shift from planktivores to benthic feeders.(wrong)}
\end{itemize}

The main reason for a wrong classification is that - although the modal verb only has influenced a part of the whole sentence - the whole sentence has been assigned as to be epistemic. Especially, composed sentences like

\begin{itemize}
    \item[$\rightarrow$]{\sf EPISTEMIC-DEONTIC  This uncertainty {\bf may} relate to the fact that Buddenbrockia genes 
have undergone rapid sequence evolution, which {\bf can} either cause artifactual groupings or reduce the support for the correct grouping. }
\end{itemize}

have been classified twice, i.e., being epistemic and deontic. This is wrong as only the first part ({\sf may}) is epistemic, the second part deontic ({\sf can}).

\section{Conclusions}
The calculation process is characterised and influences by a multitude of external contributions, therefore, one of the next steps will be a step-by-step automatisation and the access to extended sources.

Although the classification result show good results, a more detailed consideration of modal verbs may become concerned as some of them negatively and positively influence propositional sentences. Last, the lexical environment must be considered if we want to automate the general hypothesis of being the modal part is $M$ or $\neg M$. If a modal verb is discovered in the sentence structure, we can assume that the meaning is either positive or negative; it can be modified, if negations occur.

We still have in mind to constitute the modality as one possible method to characterise the author's attitude. This may be accomplished by other works of the group, i.e., the zoning of textual documents, the imaging of texts to self-organizing maps, and the fingerprinting of texts using statistic and linguistic variables. 

\section{Acknowledgement}

This work has been performed within the research project \textit{TRIAS}, which is funded by the University of Luxembourg.

\end{document}